# Towards a Cyber Information Ontology

David Limbaugh[1], Mark Jensen[1,2], and John Beverley[1,2,3,4*]

[1] *National Center for Ontological Research*
[2] *U.S. Customs & Border Protection*
[3] *University at Buffalo, Department of Philosophy, Buffalo, NY, USA*
[4] *Institute for Artificial Intelligence and Data Science*

**Abstract**
This paper introduces a set of ontology terms that are intended to act as an interface between cyber ontologies, such as a file system ontology or a data fusion ontology, and top- and mid-level Basic Formal Ontology and Common Core Ontologies. These terms center on what makes cyber information management unique: numerous acts of copying items of information, the aggregates of copies that result from those acts, and the faithful members of those aggregates that represent all other members.

## 1. Introduction

We understand 'cyber information' to refer to information carried by electronic devices, anything from copper wire to display monitors to hard drives. The purpose of this paper is to describe a set of general terms and semantics for representing cyber information in ontologies that extend from Basic Formal Ontology (BFO), Relations Ontology (RO), and the Common Core Ontologies (CCO) [1][2][3].

Two caveats regarding the scope of this paper. First, the terms introduced here are too general and too few to comprise an ontology of their own. Regardless of whether they are added to CCO, one of its extensions, or land elsewhere, these terms are a step towards allowing CCO to interface with ontologies dealing with cyber information and its management. Hence, we refer to these terms collectively as Towards a Cyber Information Ontology (TACIO)[1]. Second, there is nothing fundamentally different about cyber information compared to pen-and-paper information. The difference, rather, is found in the *management* of these respective types of information. The backbone of cyber information management is the rapid copying of information, which is found at all levels of the cyber domain [4][5]. Compare: you write a letter and mail that letter to a recipient;

---

[1] Ontology files: https://github.com/CommonCoreOntology/TowardsACyberInformationOntology

you write an email, and that email is copied, that copy is copied, and so on until a copy arrives on the intended recipient's machine. It is for this reason that we focus here on terms for the copying, duplicating, and pseudo copying of information.

Why build an interface for cyber ontologies to extend from CCO? As an extension of the BFO/CCO suite, a cyber information ontology gains interoperability with ontologies spanning medicine, biology, manufacturing, defense, and intelligence. Such interoperability is of high value given the importance and ubiquity of cyber information. We leave it to electrical engineers, physicists, and other relevant subject matter experts to make use of this interface by building more specific ontologies, like an ontology of file systems, an ontology of over-the-internet collaboration, or data fusion pipelines.

## 2. Background

Transmission of information across cyber information systems will be better understood, modeled with higher precision, and engender greater trust among users, if the rather opaque nature of the domain were adequately represented in widely used ontologies such as BFO and CCO. While at least one ontology exists that is designed to represent the domain of data fusion [6], none do so by extending from a modular suite of upper- and mid-level ontologies. Ontologies that do not conform to standardized architecture are limited in their interoperability, scalability, and ability to aid information extrapolation from diverse data sets. Crucially, as more ontologies are developed that use the same upper-levels, and themselves reused in other applications, there occurs a community-driven stepwise improvement of the reused ontologies. Ontologies that facilitate interoperability among all-source data may provide valuable machine-learning assessments of iterative data fusion pipelines, whereby slices of a pipeline are evaluated to discover how best to compose optimal pipelines, as evidenced in intelligence analysis approaches [7]. It is challenging to see how such assessment might be conducted without interoperability.

A concrete example of the need for TACIO is the current state of The Common Core Cyber Ontology (C3O) [8]. C3O is designed to represent entities relevant to the cyber domain [9]. It contains a mix of high-level terms, such as Computer Program, Computer Network, Computer Language, Digital File, as well a mix of more specific terms, e.g., Packet Header, Ethernet Cord, and Botnet. While some C3O terms are relevant to information processing and transfer, like Data Synchronization Utility and File Compressor, none explicitly characterize information tracing and authenticity validation through transformations, encodings, and manipulations of data over their digital lifespan. There is, moreover, no unifying set of terms and semantics available to C3O for representing different kinds of encoding in the cyber domain. Thus, while C3O is in general expressively rich, coverage could be improved. We intend TACIO to supplement C3O, provide motivation for refining C3O terms, and support C3O extensions, especially fusion-related applications.

Our work is guided by the CCO approach to modeling information.[2] CCO's approach to information distinguishes the content of information both from the artifacts which carry it and from the patterns exhibited by those artifacts. The screen of a computer monitor, for example, is an artifact that bears patterns – patterns of shapes and colors – that concretize information content. These distinctions allow the flexible relationship between artifacts, patterns, and information content to be represented. Consider, your monitor screen could bear any of the following distinct patterns: 'π', 'pi', '3.14...', or '3.14159265358979323...', and all of these would concretize (and convey) the same information content, assuming certain conventions in the language of contemporary mathematics. Similarly, an iPad screen might exhibit smaller patterns which also carry (and convey) the same information. Likewise, an assistive device could concretize (and convey) the same information content, using acoustic patterns to convey what is also carried by the patterns on screen.

---

[2] See documentation at: https://github.com/CommonCoreOntology/CommonCoreOntologies [3].

CCO defines the class Information Content Entity (ICE) to characterize content carried by instances of the class Information Bearing Entity (IBE) and employs BFO's class Specifically Dependent Continuant (SDC) to characterize patterns, such as shapes and colors, found in instances of IBE. SDCs are said to concretize ICEs and inhere in IBEs, while the latter are said to carry ICEs. For example:

- The patterns 'pi' and '3.14159265358979323...' that appear on your monitor screen are instances of SDC, that inhere in the screen, and concretize an Information Content Entity, call it: "ice 1"
- Your monitor screen is an instance of Information Bearing Entity that carries ice 1 which is concretized by 'π', 'pi', and '3.14159265358979323...'
- Similarly, the 'π', 'pi', and '3.14159265358979323...' that appear on your iPad screen are all instances of SDC, inhere in the screen, and concretize ice 1
- Your iPad screen is an instance of Information Bearing Entity that carries ice 1 which is concretized by 'π', 'pi', and '3.14159265358979323...'

CCO's approach to modeling information provides terminology needed to make explicit relationships among the bearers of information, the shapes encoding information, and the information itself.

Worth noting is that the carrier/content distinction implies that representing something as being concretized by 'pi' requires reference to an instance of Information Bearing Entity with the appropriate features. That is, CCO's information model rejects the view that information transmission, processing, etc., can be adequately represented without also representing carriers of that information. Carriers are necessary to track when modeling the provenance and pedigree of data, threat analysis, and data fusion across, for example, multi-modal sensors. Additionally, the clarity provided by CCO's information model allows for explication of information propagation processes, such as sending emails, duplicating files, editing files, etc., and information protection strategies, such as filesystem snapshots, encryption, and so on. These processes and their participants are at times opaque, owing to the complex relationships among the relevant cyber information artifacts, patterns, and content involved.

## 3. Methods

*3.1. Imports*

Our starting point was the top-level ontology BFO and the mid-level CCO suite. BFO was imported in full, but not all modules in the CCO suite were needed. Additionally, an extension of CCO, the Cognitive Process Ontology (CPO), was imported [10]. CPO was built from prior ontological work [11][12] and provides terms useful for modeling mental processing [9]. The latter has been used to model intelligence analyst workflows and processes of forming and asserting trustworthy medical diagnoses [8][13]. Characterization of trustworthy transmission of information among analysts and physicians is a special case of trustworthy transmission of information generally. As the latter is within the scope of cyber information, and more specifically of data fusion, domains, importing CPO is warranted for our project. On the other hand, the C3O was not imported.

CCO's Information Entity Ontology (IEO) was imported in full, as it includes terms required to express CCO's theory of information. IEO is dependent on other ontologies in the CCO suite, which were imported. The complete list of imports is thus:[3] Basic Formal Ontology (BFO), Relations Ontology (RO), Information Entity Ontology (IEO), Cognitive Process Ontology (CPO), Agent Ontology (AgO), Artifact Ontology (ArO), Extended Relation Ontology ERO). Every root term in TACIO extends from a leaf term found in one or other imported ontology. This downward population strategy facilitates defining content using an Aristotelian schema. That is, a given class 'A' is defined according to the pattern "is a B that Cs" where 'B' is the parent class of 'A', and 'Cs' designates features of 'A' that distinguish it from other instances of 'B'.

Consultation with subject-matter experts resulted in the curation of an initial term list, as well as competency questions which were ultimately used to evaluate the implementation of terminological content and identify gaps in coverage. If, for example, a competency question could not be straightforwardly represented in the query language SPARQL, then terminological content was added allowing such representation. On the other hand, assuming a toy dataset, if SPARQL queries representing competency questions returned inaccurate or unexpected results, this was taken as a reason to investigate and perhaps revise relationships among our proposed terminological contents.

Two experiments were conducted along the preceding lines. In the first, we observed that plausible competency questions could not be represented adequately given the resources of BFO, CCO, and CPO. The results prompted the introduction of additional terminological content, after which a second experiment was conducted using the same competency questions, resulting in competency questions being both representable and returning expected results.

*3.2. Competency Questions*

Competency questions provide a foundation for evaluating expressivity of the terminological content and aid in evaluating the intended scope of the ontology. Our competency questions, including examples, which drive our project are:

1. What entities are involved in the spread of information from one information bearer to another?
    a. An email is sent from a laptop to a personal computer.
    b. Information stored on a solid-state drive is displayed on a monitor.
    c. A file system snapshot is stored on a backup drive.
2. What entities are involved in the spread of information from multi-modal information bearers to a target?
    a. Acoustic and image traffic data are sent from field sensors to a controller which adjusts the timing of traffic light changes.

---

[3] Other than BFO [1], each of these can be found at https://github.com/CommonCoreOntology [3].

b. Bathymetric, meteorological, and temperature data are sent to a dataset for tracking marine life.
3. What relationship exists among information bearers participating in the spread of information from a source?
   a. Relationships among 10 machines that receive an email from a single machine.
   b. Relationships among distinct drives storing identical snapshots of a system.
4. What entities are involved in spreading information according to standard transmission protocols?
   a. A password is submitted over an encrypted Secure Socket Layer connection.
   b. A password is submitted over an unencrypted HTTP connection.
5. What relationships exist among information entities under version control which correspond to a single, current, version?
   a. Two authors collaborating on a paper using Google Docs.
   b. A developer accessing code on a GitHub repository, from distinct machines.

## 4. Results

The preceding competency questions were used to guide the use of terms from imported ontologies during our first experiment. After observing gaps in the expressivity of our terminology, we introduced additional terminological content needed to adequately answer the competency questions. The competency questions were then used to validate our results in a second experiment.

The top-level terminological content found in **Table 1** was observed to be sufficient to model, at a high level, information content, the carriers of that content, and the features that encode that content, and in this regard, was found suitable to act as an interface between the BFO/CCO suite and a cyber ontology. However, these terms, though necessary, were insufficient to model managing and transmitting cyber information.

**Table 1**
Prior Existing Terminological Content

| Label | Definition/Elucidation |
|---|---|
| *Information Bearing Artifact* | An Artifact that carries an Information Content Entity and is designed to do so using a particular format or structure. (ArO) |
| *Information Medium Artifact* | An Artifact that is designed to have some Information Bearing Artifact as part. (ArO) |
| *Information Bearing Entity* | An Object upon which an Information Content Entity generically depends. (IEO) |
| *Information Content Entity* | A Generically Dependent Continuant that generically depends on some Information Bearing Entity and stands in relation of aboutness to some Entity. (IEO) |
| *Generically Dependent Continuant* | A continuant that generically depends on one or more other entities. (BFO) |
| *Specifically Dependent Continuant* | A continuant b such that there is some independent continuant c which is not a spatial region and which is such that b s-depends on c at every time t during the course of b's existence. (BFO) |
| *has part* | A core relation that holds between a whole and its part. (BFO) |
| *g-depends on* | b g-depends on c = Def. there inheres in c a specifically dependent continuant which concretizes b. (BFO) |
| *concretizes* | A relationship between a specifically dependent continuant and a generically dependent continuant, in which the generically dependent continuant depends on some independent continuant in virtue of the fact that the specifically dependent continuant also depends on that same independent continuant. (RO) |
| *is carrier of* | b is carrier of c = Def. c g-depends on b. (RO) |
| *agent in* | x agent in y =Def y is an instance of Process and x is an instance of Agent, such that x is causally active in y. (AgO) |
| *Intentional Act* | An Act in which at least one Agent plays a causative role and which is prescribed by some Directive Information Content Entity held by at least one of the Agents. (AgO) |
| *Information Processing Artifact* | An Artifact that is designed to use algorithms to transform some Information Content Entity into another Information Content Entity. (ArO) |
| *participates in* | A relation between a continuant and a process, in which the continuant is somehow involved in the process. (RO) |
| *has input* | y has input x =Def x is an instance of Continuant and y is an instance of Process, such that the presence of x at the beginning of y is a necessary condition for the start of y. (ERO) |
| *has output* | y has output x =Def x is an instance of Continuant and y is an instance of Process, such that the presence of x at the end of y is a necessary condition for the completion of y. (ERO) |
| *Process of Proper Functioning* | A process that has been successfully vetted or designed to reliably, in environments of given types, form outputs that meet some standard and is occurring in an instance of such an environment. (CPO) |

By expanding our terms list to include all terms from the imported ontologies, we were able to represent the complexities of the items of information referred to in the competency questions. For example, the imported terms were insufficient to represent the transmission, aggregation, and trustworthiness of those items. These observations motivate the introduction of new terminological content, spanning three categories: Acts of Encoding, Aggregates of Information Content Entities, and Canonical Copies of An Information Content Entity. This new content is found in **Table 2**.

**Table 2**
New Terminological Content

| Label | Definition/Elucidation |
|---|---|
| *Act of Encoding* | An Intentional Act whereby an Agent forms a Material Entity to carry some intended Information Content Entity. (TACIO) |
| *Reference Carrier* | A material entity that is the input to an Act of Copying and that carries the ICE intended to be carried by the output of that Act of Copying. (TACIO) |
| *Act of Copying* | An Act of Encoding whereby an Agent forms a Material Entity with the intention that this Material Entity carry the same ICE as some Reference Carrier. (TACIO) |
| *Act of Information Carrier Transition* | An Act of Copying whereby an Agent forms a Material Entity that is of a type different from the Reference Carrier, but which bears the same type of concretizing SDCs. (TACIO) |
| *Act of Concretizer Transition* | An Act of Copying whereby an Agent forms a Material Entity that bears concretizing SDCs of types distinct from those inhering in the Reference Carrier, but where that Material Entity is of the same type. (TACIO) |
| *Act of Carrier and Concretizer Transition* | An Act of Copying that is both an Act of Information Carrier Transition and an Act of Concretizer Transition. (TACIO) |
| *Act of Duplication* | An Act of Copying whereby an Agent forms a Material Entity that is of the same type and bears the same type of concretizing SDCs as the Reference Carrier. (TACIO) |
| *Information Carrier Structure Entity* | A Directive Information Content Entity that prescribes the formation of a Material Entity that is a copy of a Reference Carrier. (TACIO) |
| *Act of Encoding an Information Carrier Structure Entity* | An Act of Encoding whereby an Agent forms an Information Carrier Structure Entity. (TACIO) |
| *Aggregate of Information Carrier Copies* | An Object Aggregate whose members consist of a Reference Carrier and the information descendant copies of that Reference Carrier. (TACIO) |
| *Aggregate of Duplicate Information Carriers* | An Aggregate of Information Carrier Copies all of whose members belong to the aggregate only because of one or more Acts of Duplication. (TACIO) |
| *Aggregate of Pseudo-Duplicate Information Carriers* | An Aggregate of Information Carrier Copies whose members belong to the aggregate not only because of one or more Acts of Duplication. (TACIO) |
| *has information descendant copy* | x has information descendant copy y =Def x is a Reference Carrier that participates in Act of Copying that has output y. (TACIO) |
| *has information ancestor copy* | x has information ancestor copy y =Def y has information descendant copy x. (TACIO) |
| *is a canonical copy of* | x is a canonical copy of y =Def x is the output of an Act of Copying that is a Process of Proper Functioning, y was the Reference Carrier in that Act of Copying, and there is no sufficient reason to doubt the faithfulness of x as a copy. (TACIO) |
| *has canonical copy* | x has canonical copy y =Def y is a canonical copy of x. (TACIO) |
| *is a canonical member of* | x is a canonical member of y =Def x is the output of an Act of Copying that is a Process of Proper Functioning, and the Reference Carrier of x is the earliest ancestor of Aggregate of Information Carriers y. (TACIO) |
| *has canonical member* | x has canonical member y =Def y is a canonical member of x. (TACIO) |

## 5. Discussion

*5.1. Acts of Encoding*

Cyber information systems spread ICEs by creating copies of IBEs which carry the same content. For example, emailing involves using a sender's IBE as reference in a copying process which produces a distinct IBE on the recipient's machine. If successful, each IBE carries the same ICE.

To capture these phenomena, we first introduce the term Act of Encoding: An Intentional Act whereby an Agent forms a Material Entity to carry some intended Information Content Entity. As an intentional act, any instance of Act of Encoding involves an agent intentionally arranging, manipulating, or creating some material entity for the purpose of carrying specific information content. This class is broad enough to include writing in a notebook, using fireworks to display a message, as well as an agent using a computer to manipulate magnetic fields to store information. This term thus accommodates ways in which cyber information systems manipulate IBEs to store, protect, or transmit ICEs.

We moreover introduce a subclass of Act of Encoding, *Act of Copying*: An Act of Encoding whereby an Agent forms a Material Entity with the intention that this Material Entity carry the same ICE as some Reference Carrier. Consequently, we also introduce Reference Carrier: A material entity that is the input to an Act of Copying and that carries the ICE intended to be carried by the output of that Act of Copying. Thus, if the ICE carried by the output Material Entity is the same ICE carried by the Reference Carrier, then the Act of Copying is successful. For example, if I copy a quote from a book by hand, then part of the book is the Reference Carrier, and, if what I copy by hand carries the same ICE as the relevant part of the book, then the Act of Copying was successful. Similarly, if an ICE carried by a portion of my laptop is sent to some recipient, then this Act of Copying is successful if the process outputs a Material Entity (in this case, part of the recipient's laptop), which carries the same ICE as was carried by the relevant portion of my laptop. (Note, to be clear, a failed, or unsuccessful, act of copying is one where the output IBE carries an ICE different from the Reference Carrier. There is no such thing as an Act of Copying where there is no output.)

Of course, Acts of Copying may spread information across different carrier types, like copying notes from paper to chalk board, inputting handwritten manuscripts into a word processor, or an operating systems' copying content from persistent storage to RAM. To represent such processes, we introduce Act of Information Carrier Transition: An Act of Copying whereby an Agent forms a Material Entity that is of a type different from the Reference Carrier, but which bears the same type of concretizing SDCs.

Copying may also spread information across distinct types of concretizing SDCs. Some potential examples include changing the font type of a text, translating a text into an equally expressive language, and performing certain algebraic operations in mathematics. This motivates the introduction of Act of Concretizer Transition: An Act of Copying whereby an Agent forms a Material Entity that bears concretizing SDCs of types distinct from those inhering in the Reference Carrier, but where that Material Entity is of the same type.

There are also many cases where the carrier and the concretizers both transition. For example, consider a voice-to-text application that transitions information concretized by patterns of acoustic waves to, say, a binary pattern of electromagnetic features inhering in a machine. Hence, we introduce Act of Carrier and Concretizer Transition: An Act of Copying that is both an Act of Information Carrier Transition and an Act of Concretizer Transition. Notably, this is a defined class, characterized entirely in terms of other TACIO classes.

Some Acts of Copying spread information by duplication, where a Material Entity is produced that is exactly similar (or something relevantly close to this) as the reference carrier; in the parlance previously used, there is no transition at all. For example, a file system that creates snapshots of files, for the sake of redundancy, is likely creating duplicates in our intended sense. Thus, we introduce the class Act of Information Duplication: An Act of Copying whereby an Agent forms a Material Entity that is of the same type and bears the same type of concretizing SDCs as the Reference Carrier.

Lastly, some acts of encoding produce a specification for what an act of copying should output. Thus, though the relevant ICE is not carried by the instructions, it is reproducible, such as the processes through which data packets are fragmented then reassembled during network transmission, according to Internet Protocol. We introduce two classes to address such scenarios. Information Carrier Structure Entity: A Directive Information Content Entity that prescribes the formation of a Material Entity that is a copy of a Reference Carrier; and Act of Encoding an Information Carrier Structure Entity: An Act of Encoding whereby an Agent forms an Information Carrier Structure Entity. Each of these entities is important in modeling acts of copying in a fine-grained manner in the cyber domain.

*5.2. Aggregates of Information Carriers*

Acts of Copying result in aggregates of IBEs, which should have in common the ICEs they carry but which are linked by relations that are a product of a shared ancestry of reference carriers and acts of copying. We formalize the identification of these aggregates using the following transitive relations:

- x has information descendant copy y just in case x is a Reference Carrier in an Act of Copying that has output y
- x has information ancestor copy y just in case y has information descendant copy x

Using these relations, we define three types of information aggregates:

- Aggregate of Information Carrier Copies: An Object Aggregate whose members consist of a Reference Carrier and the information descendant copies of that Reference Carrier.
- Aggregate of Information Carrier Duplicates: An Aggregate of Information Carrier Copies all of whose members belong to the aggregate only because of one or more Acts of Duplication.
- Aggregate of Information Carrier Pseudo Duplicates: An Aggregate of Information Carrier Copies whose members belong to the aggregate not only because of one or more Acts of Duplication.

These classes and relations in hand, we have resources useful for grouping copies of cyber information according to grades of fidelity.

*5.3. Canonical Information and Faithful Copies*

Our final additions involve the representation of trust with respect to cyber information of the sort we have discussed so far. Our strategy is to approach representing trust through the lens of epistemology, derived from notions of warrant [14][15] and defeaters [16]: If x is warranted then x is to be trusted as veridical unless there are defeaters for trusting x is veridical. A defeater for trusting x is veridical is evidence that either rebuts – shows x is not veridical – or undercuts – shows reasons for trusting x are questionable.

This plausible principle has a closely related analogue relevant to acts of encoding: If x is canonical, then x is to be trusted as a faithful encoding of some intended ICE unless there are defeaters for trusting x as a faithful encoding. This notion of canonicity appears to underwrite many functions in the cyber information domain. Without canonicity, every email received, file opened, or backup restored, should be suspect. Canonicity is thus applicable to copies of information carriers, whether created by a file system or when sending an email, that can be trusted to carry intended information.

Importantly, canonicity applies to encodings, and thus to IBEs and their concretizing SDCs rather than ICEs. IBEs are what get copied, while ICEs spread when an Act of Copying is successful. Canonicity tells us which copies can be trusted as faithful to their Reference Carriers. Importantly, it does not follow from an IBE's being canonical that it is in fact a faithful copy; just as it does not follow from someone's being trustworthy that they are not lying. Canonicity can be thought of as a nominal measurement of the reliability of some copy as faithful to its reference carrier. If x is measured as canonical, then x is measured as "to be trusted as a faithful copy of its reference carrier, absent defeaters."

Following notions of warrant, we take it that a copy of an IBE has a canonical relationship with its Reference Carrier when the following conditions are met:

1. The relevant information carrier is the output of a vetted copying process designed to, in some environment, reliably output information carriers faithful to the Reference Carrier.
2. The relevant Act of Copying occurs in a vetted environment designed to facilitate the reliable output of information carriers faithful to the Reference Carriers.

An IBE that satisfies each of these conditions is the output of what CPO considers a process of proper functioning, and thus, we should trust that the IBE is a faithful copy of its reference carrier; that is, the IBE is canonical. If these conditions are not met, an IBE may in fact be a faithful copy, but we should not trust it is a faithful copy. With this discussion in mind, we provide (merely) sufficient conditions for the following relations:

- x is a canonical copy of y just in case x is the output of an Act of Copying that is a Process of Proper Functioning, y was the Reference Carrier in that Act of Copying, and there are no defeaters for trusting the faithfulness of x as a copy of y.
- x has canonical copy y just in case y is a canonical copy of x.

Furthermore, by establishing an IBE as canonical according to a Reference Carrier, we can also establish that an IBE is canonical according to an Aggregate of Information Carriers. An IBE that is a canonical member of an aggregate represents the ICE(s) that all other members should carry and, depending on the aggregate, what all other members should be like. Here are the relations:

- x is a canonical member of y just in case x is the output of an Act of Copying that is a Process of Proper Functioning, and the Reference Carrier of x is the earliest ancestor of Aggregate of Information Carriers y.
- x has canonical member y just in case y is a canonical member of x.

These relations allow for an Act of Copying to have a series of Acts of Copying as occurrent parts, and for one or more IBEs to be the canonical members of the aggregate that is formed because of that series. Consider, when you look at the file size of some document on a computer, the computer likely has numerous redundant copies of that file in storage, and the size the computer shows you is of some canonical member of the aggregate of those files. Of course, not all Acts of Copying transfer canonicity form one copy to another. Sometimes acts of copying are not successful, or a copy gets corrupted-in place, not to mention limitations of copying fidelity.

Importantly, there are likely other ways to establish some copy as canonical, we have only supplied sufficient conditions. For example, using checksums or other methods of verification apart from assessing the Act of Copying itself. Our goal here was not to be exhaustive, but to introduce the notion of canonicity and discuss its use.

## 6. Application

Here we apply the new terminological content using SPARQL to answer question 2a from the list of competency questions above:

> 2. What entities are involved in the spread of information from multi-modal information bearers to a target?
>    a. Acoustic and image traffic data are sent from a field sensors to a controller which adjusts the timing of traffic light changes.

The question concerns the spreading and fusing of information from distinct sources with distinct types of carriers and concretizers. The query demonstrates the way the new terminological content can model carriers, their information content, and the various relationships between. In this query we exploit the fact that descendent and ancestor copies share information content. This is but one way this competency question could have been answered. The query is as follows:

```
# Title: TACIO Competency Question 2a
# Description: Acoustic and image traffic data are sent from field sensors
# to a controller which adjusts the timing of traffic light changes.

PREFIX rdf: <http://www.w3.org/1999/02/22-rdf-syntax-ns#>
PREFIX cco: <http://www.ontologyrepository.com/CommonCoreOntologies/>
PREFIX obo: <http://purl.obolibrary.org/obo/>
PREFIX tacio:
<http://www.ontologyrepository.com/CommonCoreOntologies/Exp/NewInformatio
nOntology>

SELECT DISTINCT *
WHERE {
?recording_process_1 cco:has_output ?iba_1 .
?iba_1 a cco:AudioRecording ;
       tacio:has_information_descendant_copy ?iba_3 ;
       obo:RO_0010002 ?ice_1 . #is carrier of
?ice_1 cco:describes ?traffic_event_1 .

# A recording process outputs an audio recording that carries information
# content describing a traffic event, and which is a reference carrier #
# having an information bearing artifact as a descendent copy.

?recording_process_2 cco:has_output ?iba_2 .
?iba_2 a cco:Image ;
       tacio:has_information_descendant_copy ?iba_4 ;
       obo:RO_0010002 ?ice_2 . #is carrier of
?ice_2 cco:describes ?traffic_event_1 .

# A recording process outputs an image that carries  information content
# describing the same traffic event, and which is a reference carrier
# having an information bearing artifact as a descendent copy.

?controller_1 a cco:ControlSystem ;
              cco:agent_in ?act_of_timing_change_1 ;
               obo:BFO_0000051 ?processor_1 . #has part
?processor_1 a cco:InformationProcessingArtifact ;
               cco:agent_in ?act_of_processing_1 .
```

```
?act_of_processing_1 a cco:ActOfArtifactEmpoyment ;
                     cco:has_input ?iba_3, ?iba_4 ;
                     cco:has_output ?iba_5 ;
                     cco:has_process_part ?act_of_copying_1 ;
                     cco:has_process_part ?act_of_copying_2 .
?act_of_copying_1 a tacio:ActofCarrierandConcretizerTransition ;
                  cco:has_input ?iba_3 ;
                  cco:has_output ?iba_6 .
?act_of_copying_2 a tacio:ActofCarrierandConcretizerTransition ;
                  cco:has_input ?iba_4 ;
                  cco:has_output ?iba_7 .
?iba_5 cco:input_of ?act_of_timing_change_1 ;
       obo:BFO_0000051 ?iba_6 ; #has part
       obo:BFO_0000051 ?iba_7 ; #has part
       obo:RO_0010002 ?ice_3 . #is carrier of
?ice_3 cco:describes ?traffic_event_1 .
   }

# A processor part of a control system processes the descendent
# information bearer copies, resulting in an information bearer used as
# input to an act of timing change, and which carries another description
# of the traffic event.
```

## 6. Future Direction

We have taken first steps here towards modeling cyber information using high-level terms in a manner conformant to widely used top and mid-level ontologies. As described, the conformance of TACIO terminological content allows for interoperability with approximately 400 BFO-based ontologies. Because the cyber domain cuts across domains such as biology, education, manufacturing, and so on, the interoperability of TACIO with existing ontologies provides significant, wide-ranging, content of the sort that TACIO terminological content is about.

We welcome the participation of those who are experts in the cyber, defense, and intelligence domains, as we refine TACIO terminological content, and encourage contributions through the TACIO GitHub project page. Next steps include applying TACIO terminological content to modeling real-world datasets, as well as more general phenomena, such as in representing encoding errors, unanticipated duplication or transmission of information, and transfer protocols. We anticipate such modeling to result in the construction of additional competency questions and to identify gaps in TACIO coverage. Other steps going forward will include: working with developers of the CCO and C3O in determining which, if any, terms from TACIO should be curated in these respective ontologies and in deciding if any existing content in those ontologies may need to change in light of this research; and applying this work to modeling information processing artifacts and software and their respective functions, especially as it pertains to the manipulation of carriers and the extraction or creation of new content; and the tracking of pedigree and provenance across the digital lifecycle. Our refinements can only be improved through consultation with additional subject matter experts and ontologists invested in coordinating cyber information data.